\documentclass[10pt]{article}
\usepackage{amsmath, times, cite,microtype, verbatim}
\usepackage[top=1in, bottom=1in, left=1in, right=1in]{geometry}
\usepackage{amssymb}
\usepackage{amsthm}
\usepackage{amscd}
\usepackage{amsfonts}
\usepackage{graphicx}%
\usepackage{lastpage}
\usepackage{caption,tabularx}
\usepackage{enumitem, inconsolata}
\usepackage[citecolor=black,linkcolor=black,colorlinks=true,urlcolor=blue]{hyperref}

\usepackage{etoolbox}

\newcommand{\minihead}[1]{{\textbf{#1.} }}

\newcommand{\thrust}[3]{{\noindent\emph{#1 \emph{(#2)}:} #3}}

\newcommand{\fs}{feature engineering}
\newcommand{\dataprep}{data preparation}
\newcommand{\product}{productionization}

\begin{document}
\pagestyle{empty}
\thispagestyle{empty}

\begin{center}
\LARGE{Infrastructure for Usable Machine Learning:\\The Stanford DAWN Project}\\[4mm]

\large{ Peter Bailis,  Kunle Olukotun, Christopher R\'{e}, Matei Zaharia\\
Stanford DAWN Project\\
{\normalsize \url{http://dawn.cs.stanford.edu/}}\\[4mm]

21 May 2017}
\end{center}

\begin{center}
\textbf{Abstract}\vspace{-1em}
\end{center}

{Despite incredible recent advances in machine learning, building
  machine learning applications remains prohibitively time-consuming and expensive for all but the best-trained, best-funded engineering
  organizations. This expense comes not from a need for new and
  improved statistical models but instead from a lack
  of systems and tools for supporting \emph{end-to-end} machine learning application development, from data preparation and labeling to productionization and monitoring. In this document, we outline opportunities for infrastructure supporting usable, end-to-end machine learning applications in the context of the nascent DAWN (Data Analytics for What's Next) project at Stanford. }

\section{Introduction and DAWN Project Goals}

\minihead{A Gilded Dawn for Machine Learning and Artificial Intelligence} We are in the golden age of machine
learning and artificial intelligence.  Sustained algorithmic advances
coupled with the availability of massive datasets and fast parallel
computing have led to breakthroughs in applications that would have
been considered science fiction even a few years ago. Over the past
five years, voice-driven personal assistants have become commonplace,
image recognition systems have reached human quality, and autonomous
vehicles are rapidly becoming a reality.  Given these successes,
there is no doubt that machine learning will transform most areas of
our economy and society.  Businesses, governments and scientific labs are clamoring to
see how machine learning can tackle their problems.

Unfortunately, although new machine learning (ML) applications are
impressive, they are very expensive to build.  Every major new ML
product, such as Apple Siri, Amazon Alexa, or Tesla Autopilot,
requires large and costly teams of domain experts, data scientists,
data engineers, and DevOps. Even within organizations that have
successfully employed ML, ML remains a rare and expensive commodity
reserved for a small subset of teams and applications. Moreover, many
ML models require huge amounts of training data, and obtaining such
training data is highly challenging in many application domains. For
example, even though an ML algorithm may achieve human accuracy in
identifying pictures of dogs on the Internet (thanks to millions of
available labeled images), the algorithm will not achieve the same
accuracy identifying cancer in medical images unless an organization
expends years of human expert time creating labeled training data.
Finally, once an ML product is built, it requires substantial effort to
deploy, operate, and monitor at scale, especially if critical business
processes will rely on it. For example, how can a company make
guarantees about its new automated diagnosis system, or monitor that
the system is performing as expected in practice? In summary, ML
technology is at a similar stage to early digital computers, where
armies of white-clad technicians labored to keep a small handful of
machines operating in production: ML technology clearly has tremendous
potential, but today, ML-powered applications are far too expensive
to build for most domains.

\minihead{Our Response: The DAWN Stack for End-to-End ML Development}
To address this potential, our group at Stanford is beginning a new,
five-year research project to design systems infrastructure and tools
for \textit{usable machine learning}, called DAWN (Data Analytics for
What's Next).  Our goal is \emph{not} to improve ML algorithms, which
are almost always ``good enough" for many important applications, but
instead to make ML \emph{usable} so that small teams of non-ML experts
can apply ML to their problems, achieve high-quality results, and
deploy production systems that can be used in critical applications.
Whereas today's ML successes have required large and costly teams of
statisticians and engineers, we would like to make similar successes
attainable for domain experts, such as a medical lab optimizing clinical
procedures or a business group applying ML to its domain-specific
problems. Major improvements in the usability of machine learning are
mandatory to realize its potential. We ask:

\emph{How can we enable anyone with domain expertise to build their own production-quality data products \emph{(without requiring a team of PhDs in machine learning, big data, or distributed systems, and without understanding the latest hardware)}?}

At first, our goal of usable machine learning might appear too
ambitious---how can we expect one or two domain experts to match work
that today requires teams of hundreds?  Our observation is that such
revolutions ``democratizing'' computing technology have happened
before.  For example, although textual search is a complex field
requiring sophisticated algorithms and data structures, today, search
is ubiquitous. Non-expert users rely on search engines every day, and
any developer can add search to an application by linking a library
such as Lucene or Solr.  These libraries offer good enough results out
of the box---as well as simple enough tuning options---to be usable by
non-experts.  Similarly, in the 1970s, relational databases
revolutionized data management.  Before these modern databases,
organizations built computer applications using low-level code that
had to directly manipulate on-disk data structures and implement
complex processing algorithms.  Databases encapsulated this complexity
behind simple interfaces that any developer can use, and that most
users can even \emph{tune} without understanding system internals.  As
a result, organizations need to spend far less effort to build a
data-intensive application, and, instead of running one or two custom,
legacy database-backed applications, many organizations run thousands
of database-backed applications every day.

With history as a guide, our key observation is that most of the
effort in industrial ML applications is \emph{not} spent in devising
new learning algorithms or models but is instead spent in other areas
that are especially in need of better tools and infrastructure:
\emph{data preparation}, \emph{feature selection and extraction}, and
\emph{productionization} (cf. ~\cite{ml-credit-card}).  Data
preparation means acquiring, producing and cleaning enough training
data to feed into an ML algorithm: without this quality data, ML
algorithms fall flat. Feature selection and extraction means
identifying the data characteristics and behaviors of interest: what
aspects of data are most important, and what would a domain expert
implicitly or explicitly say about a given data point?
Productionization means deploying, monitoring and debugging a robust
product: how can an organization check that the ML algorithm deployed
is working, debug issues that arise, and make the system robust to
changes in data?  In the large teams that build ML products such as
Siri, most of the individuals work on data preparation, feature
selection and extraction, and productionization, as well as the
distributed systems infrastructure to drive these tasks at scale,
\emph{not} on training ML models. However, thus far, these critical
steps in the ML product pipeline have received far less attention than
model training and new model tweaks---both from the research community
and the open source software community---and, based on our prior work
in this area, we see substantial opportunity to greatly reduce the
effort required by these tasks via the development of new software
tools and systems infrastructure.

\section{The DAWN Project and Systems Research in Usable ML}

\begin{figure}
\begin{center}
\includegraphics[width=.8\columnwidth]{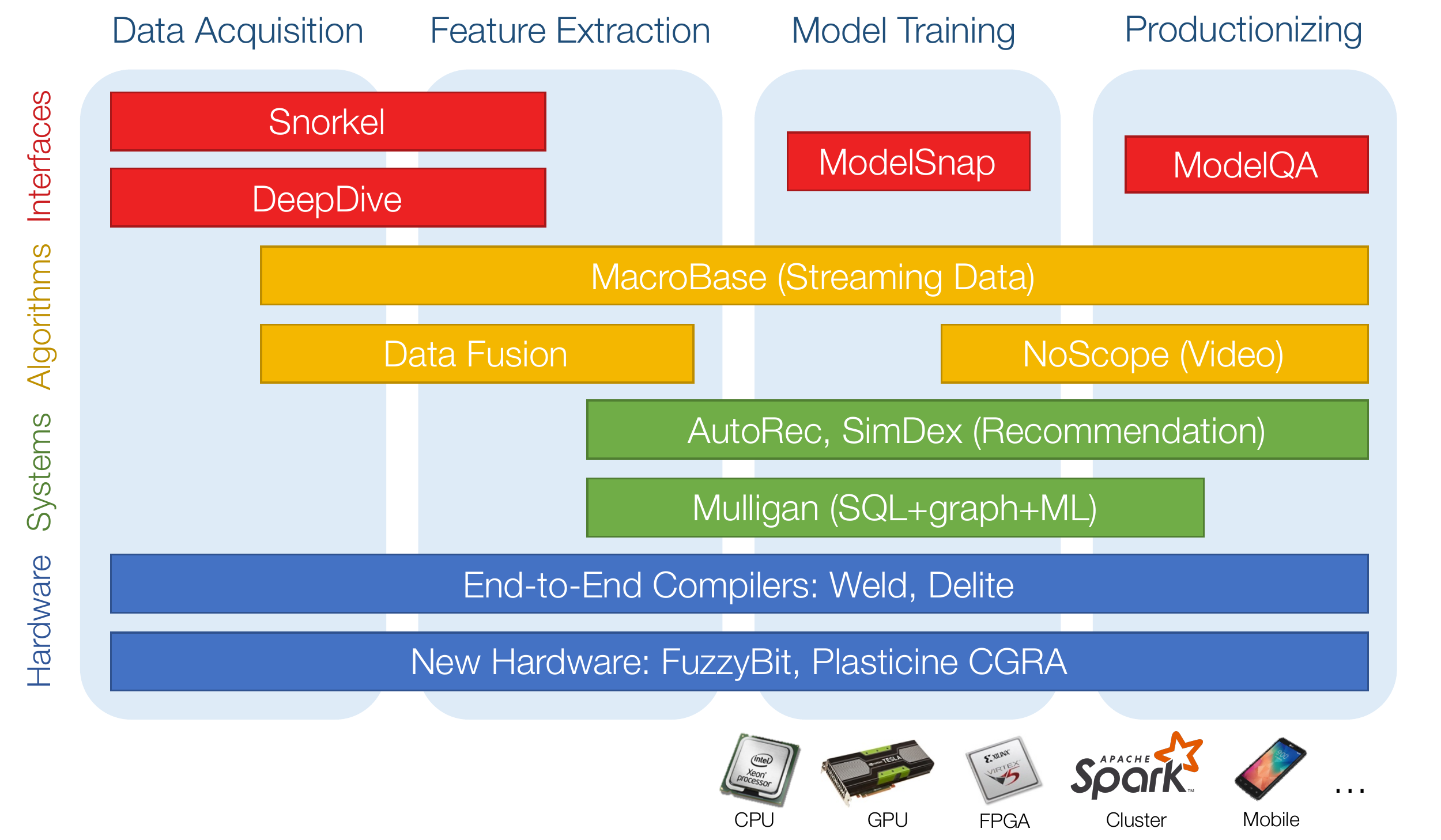}\vspace{-1em}
\end{center}
\caption{The DAWN Stack for Usable Machine Learning: In the Stanford DAWN project, we are addressing the need for infrastructure for usable ML by building a research stack of software and tools spanning each stage of the ML lifecycle and abstractions from new interfaces to new hardware. We believe this parallel end-to-end and interfaces-to-hardware approach is necessary to fully realize the potential of more usable ML.}
\label{fig:stack}
\end{figure}

To capitalize on the opportunity represented by our project goals and
drawing on our prior experience in building large-scale analytics
systems such as Apache Spark~\cite{spark}, Apache Mesos~\cite{mesos},
Delite~\cite{delite} and DeepDive~\cite{deepdive}, we are spending the
next five years researching and building tools to address these end-to-end
problems in usable machine learning. By combining expertise spanning
algorithms to systems to hardware, working closely with a small team of
collaborative partners in some of the most challenging data-intensive
domains, and producing and validating high-quality research
prototypes in real application scenarios, we plan to tackle DAWN's
goals at multiple stages of the ML lifecycle and levels of the systems
stack (see Figure~\ref{fig:stack}).

Our design philosophy in the DAWN
stack centers around three main tenets:

\textbf{a) Target end-to-end ML workflows.} ML-powered application
development consists of far more than model training. As a result, today,
the bulk of challenges in developing new ML-powered applications are
\emph{not} in model training but are instead in data preparation,
feature selection/extraction, and productionization (serving, monitoring,
debugging, etc). Systems should target the entire, end-to-end ML workflow.

\textbf{b) Empower domain experts.} The highest-impact ML applications will have to be developed by domain experts, not ML experts. However, today, few systems allow these domain experts to encode their domain knowledge so it can be leveraged via automation and machine learning models. Systems should empower users who are not ML experts, by providing them tools for onerous tasks such as labeling, feature engineering and data augmentation.

\textbf{c) Optimize end-to-end.} Execution speed is important in ML both for
model training, where it allows building \emph{better} models (e.g., through
more input data or wider parameter search), and for production serving, where
it allows deploying these models cost-effectively in practice.
However, today's ML tools often perform 10--100$\times$ below hardware limits,
requiring expensive software engineering to build production systems.
Our early results show that by architecting tools to optimize ML pipelines
end-to-end, and leveraging statistical properties of the algorithms such
as tolerance to inexact execution, we can accelerate ML applications by
10--100$\times$ on both current and emerging hardware.

In summary, we believe that systems that target full application development, real non-expert needs, and optimize all levels of the software and hardware stack are critical to fully realizing the potential of usable ML.\vspace{1em}

\subsection*{DAWN Research Directions} To embody these principles, we are pursuing research along several directions. We provide an overview below and provide citations to early results in each area:

\textbf{I) New Interfaces to ML.} To empower domain experts who are
not ML experts, we need to develop new interfaces to ML technologies,
from model specification to model monitoring:

\begin{enumerate}[itemsep=.1em,parsep=.4em,topsep=.5em, label={\alph*)}]
\item \thrust{Easing model specification via observational ML}{\dataprep, \fs}{
Can we build ML systems that learn high-quality models simply by observing domain experts? For example, when labeling data, domain experts often apply a set of heuristic rules to determine the label for a given data point (e.g., if the phrase ``preheat the oven'' appears repeatedly in a document collection, label the collection as likely pertaining to cooking). By providing simple interfaces for these users to specify their beliefs about data in rule form (e.g., regular expressions), we can combine a small number of these rules and apply them to massive datasets. We use unsupervised ML to denoise the rules and learn their accuracies, and train supervised ML models with the resulting probabilistic labels, in a new paradigm we call data programming~\cite{dataprogramming}. We have obtained promising early results with a new system called Snorkel~\cite{snorkel} that produces high-quality models from low-quality rules. We are also pursuing new lines of research in weakly supervised ML to improve model quality without manual user intervention, such as feature discovery~\cite{socratic,flipper} and structure learning~\cite{bach-structure}.}

\item \thrust{Explaining results to humans}{\fs, \product}{Given an ML deployment, how can we explain ML model results to humans? As models are used in increasingly business-critical applications, the ability to explain the prediction of a classification decision in a human-interpretable manner is critical. This is challenging: large, complex models deliver highly accurate results but are far from interpretable or explainable. One promising approach is that ML predictions are not made in a vacuum: each user has tens to hundreds of attributes that can be used to segment, correlate, and contextualize predictions (e.g., users running version v47 of the software are abnormally likely to be flagged as spammers). Preliminary results with even basic correlation-based analyses~\cite{macrobase-cidr} have been extremely promising, and we plan to expand this suite of functionality to other domains, including textual, visual, and time-series data~\cite{asap}. }

\item \thrust{Debugging and observability}{\fs, \product}{ML model ``drift,'' in which phenomena evolve but models do not, can be catastrophic: for example, Google Flu Trends, which used common search terms as a signal for influenza prevalence, was prominently featured in a 2008 \emph{Nature} paper, only to later miss the peak of the 2013 flu season by an extremely large margin~\cite{flu}. As ML models are deployed, they must be monitored and updated. We are interested in developing and deploying inexpensive, useful tools for monitoring the quality of ML model predictions, especially as new models are released to potentially heterogeneous user and device platforms. Subsequently surfacing and correcting for deviations from expected behavior will require advances in both interfaces and model training.  }

\item \thrust{Assessing and enriching data quality}{\dataprep, \fs}{High-quality models are produced by consuming and training on a diverse diet of high-quality data. As more and more data sources are captured in digital form, integrating structured (e.g., data warehouse, CSV) and unstructured (e.g., text, image, and time-series) data will become increasingly important to extract signal in model construction. Given a menu of diverse data sources, which sources can be most trusted? Which sources should be augmented and enriched, either via additional human labeling or by  augmentation with existing knowledge bases? Our early results~\cite{slimfast} indicate that, if we start to \textit{explicitly model} the quality of each data source, then we can automatically identify the data sources that are most in need of enrichment, thus reducing the cost of data cleaning and acquisition. }

\end{enumerate}

\textbf{II) End-to-End ML Systems.} We believe that in many important domains, it is possible to design end-to-end systems that encapsulate the whole ML workflow and hide internals from users, similar to a search engine or a SQL database. We are pursuing several such areas:

\begin{enumerate}[itemsep=.1em,parsep=.4em,topsep=.5em, label={\alph*)}]
\item\thrust{Classification over massive streams}{\dataprep, \fs, \product}{Classification and ranking are core operators behind every modern search engine. However, how can we go beyond classifying static text or images in batch and start to classify sensor data, time series, and other data streams as they change, in real-time and at scales of tens of millions of events per second? We are interested in developing high-quality but extremely optimized operators for classification and aggregation of diverse data, combining feature transformation, classification, and aggregation over streams. Preliminary prototyping in the MacroBase engine~\cite{macrobase-sigmod} has revealed that a small number of operators can be reused at scale across domains including sensors from manufacturing, mobile analytics, and automotives. We are interested in expanding this functionality to domains such as video processing, where a \$0.50 image sensor currently requires a \$1200 graphics card to process in real time; exploiting classic systems techniques including caching, incremental memoization, branch-and-bound pruning, and adaptive specialization (e.g., training a scene-specific object detector) within a unified systems framework and ``classifier toolbox'' will enable line speed without compromising accuracy~\cite{noscope,tkde}. }

\item \thrust{Personalized recommendations}{\fs, \product}{Personalization is key to many popular ML-powered applications, and the literature is replete with algorithms for personalized recommendation. However, despite the simple inputs and outputs to recommendation engines, practitioners still have to build each engine from scratch, chaining together low-level algorithms and tools. We plan to build a general end-to-end platform for recommendation, including a simple interface for inputs (e.g., clicks or ratings from users), automatic model tuning, and automatic serving, monitoring, and model retraining. Our early results suggest that it is possible to perform all these tasks incrementally as inputs arrive, creating a ``plug-and-play'' personalized recommendation system where users can simply input user interactions and request up-to-date recommendations in real time.}

\item \thrust{Combining inference and actuation}{\fs, \dataprep, \product}{If ML is powerful because it delivers broader insights and decision-making ability, then how do we actually automate the decision-making process? Today, this combination of inference/prediction (i.e., predicting what will occur) and actuation/decision-making (i.e., taking action based on a prediction) is almost always performed by separate systems (often an automated inference engine and a human ``decider''), except in a small handful of applications such as autonomous vehicles. How do we integrate actuation and decision-making as a first-class citizen in ML pipelines? With the advent of the automated API, decision-making has never been easier (e.g., send a POST request to an automated control center); what is missing is the ``glue'' required to integrate ML and these automated APIs as well as the logic for reasoning about the combination. We are developing a series of integrations for this kind of actuated inference, from alerting and notifications to physical manipulation of the environment (e.g., send an Internet-powered Roomba to verify the presence of a student in the office).}

\item \thrust{Unifying SQL, Graphs, and Linear Algebra}{\product}{ML product pipelines consist of a diverse set of operations including, SQL, graph computations, and ML training and evaluation. Unfortunately, most execution engines optimize for only one of these computational patterns; how can we build an engine that is optimized for each of them? Perhaps surprisingly, many of these patterns can be cast as an instance of the classic relational \textit{join} operator, and PI R{\'e} recently developed an asymptotically faster join operator~\cite{join}. In practice, we have found that, when combined with SIMD-optimized execution this optimized join is fast, matching optimized engines for each of SQL and graphs~\cite{emptyheaded}. What about ML? We believe it is possible to we can do the same for many ML workloads, by extending these theoretical results to classic ML patterns including linear algebra operations and sparse matrix operations~\cite{matrixtheory}. By combining these operators within a single engine, we can optimize end-to-end pipelines of SQL, graph computations, linear algebra, and more.}

\end{enumerate}

\textbf{III) New Substrates for ML.} Training and deploying ML quickly and in a cost-effective manner requires the development of new computational substrates, from language support to distributed runtimes and accelerated hardware.

\begin{enumerate}[itemsep=.1em,parsep=.4em,topsep=.5em, label={\alph*)}]

\item \thrust{Compilers for end-to-end optimization}{\fs,\product}{Modern
ML applications are comprised of an increasingly diverse mix of libraries and systems such as TensorFlow, Apache Spark, scikit-learn, and Pandas. Even if each of these libraries is optimized in isolation, real pipelines \emph{combine} multiple libraries, so production use at scale usually requires a software engineering team to rewrite the whole application in low-level code. We are developing Weld~\cite{weld}, a new runtime that can optimize data-intensive code \emph{across} different libraries and functions to automatically generate fast implementations either for ML training or serving. Perhaps surprisingly, Weld can already accelerate modern data analysis tools such as Apache Spark, Pandas and TensorFlow by 10$\times$ by optimizing across the operators within them, and can accelerate cross-library workloads by up to 30$\times$. Moreover, Weld is designed for portability to heterogeneous hardware; we will therefore also be able to run these libraries on GPUs, mobile processors, or FPGAs. Apart from Weld, we are also developing new compiler technology for ML in Delite~\cite{delite}, a framework for developing domain-specific languages, and Splinter~\cite{splinter}, a privacy-preserving data analysis platform.}

\item \thrust{Reduced precision and inexact processing}{\product}{ML operators are stochastic and probabilistic; how can we leverage this fact in our execution substrates? Our earlier work (HogWild!~\cite{hogwild}) was the first to show that asynchrony in execution can actually improve convergence time, and the basic algorithms are now running daily in production at companies including Google, Microsoft, and other large-scale technology companies. However, we believe it is possible to go even further, lowering power and increasing performance by leveraging stochasticity at the \textit{bit level}: we can designing chips that are specialized for ML, operating at lower precision and allowing fabrication at high yield and execution at extremely low power. Our recent theoretical results illustrate low-precision execution is possible without compromising accuracy~\cite{buckwild}, with promising results in practice~\cite{buckwild-isca}.}

\item \thrust{Reconfigurable hardware for core kernels}{\fs, \product}{Computer architects commonly proclaim that year $N$+1 is the year of the FPGA. However, FPGAs remain notoriously difficult to program and expensive to utilize. Nevertheless, ML may be a turning point: in 2017, compute is an increasingly critical bottleneck for data-hungry ML analyses, both at training time and inference time. Given the impending collision of CPUs and on-chip FPGAs, reconfigurable hardware with high-level programmability functionality will be increasingly important. In addition, we are developing new substrates in the form of reconfigurable architectures for easily specifying modular and efficient compute kernels~\cite{plasticine} that will be critical to realizing performance-per-watt, especially as the upper levels of the software stack continue to evolve.}

\item \thrust{Distributed runtimes}{\product}{As models continue to grow, scale-out execution of training and inference is becoming increasingly important. Combining ML with distributed systems is a real headache: is a model misbehaving because it is distributed to too many servers, or because it is poorly specified? What's the optimal amount of asynchrony? What does the optimal distributed training framework really look like? We are extremely interested in harnessing both intra-device (e.g., FPGA, GPU, vectorized) and inter-device (e.g., cluster compute) parallelism to consume all possible resources (i.e., automatically and dynamically offloading to different hardware within a cluster). And, perhaps surprisingly, some of our recent theory~\cite{omnivore} shows that we can explicitly and automatically \textit{tune} the underlying learning algorithms for optimal execution on a given set of hardware and a given computer network. There are many remaining questions here: how can distributed asynchronous execution benefit us at inference time (i.e., in model serving)? Can we leverage new computational substrates like serverless computing (e.g., Amazon Lambda) to further scale-out inferential procedures? What is the unified programming model for distributed execution? We plan to build tools (and integrate with existing frameworks such as TensorFlow and Spark) to answer these questions.  }
\end{enumerate}

\subsection*{Research Roadmap and Success Metrics}

The DAWN research roadmap represents an exciting potential for the
future of systems and ML research and practice. Within DAWN, we will
pursue the above research objectives in collaboration with target
research partners and on- and off-campus use cases. Our primary
success metric will be usability, comprising $i)$ the time and cost to
specify an ML application (including data sources and features of
interest), $ii)$ the time and cost to execute the application in
production (including hardware and human resources to monitor the ML
models), and $iii.)$ the benefit to the end-user expert. We intend to
make all our work available as open source software, enabling
practitioners throughout industry and science to try our ideas and
benefit from DAWN's successes.

Our focus on end-to-end real-world problems lends itself naturally to
integration between components of the DAWN project stack. Simply
solving one problem, such as observational ML, without accounting for
the hardware costs at the opposite end of the stack, will lead to
sub-optimal results according to the above metrics (especially $ii)$)
in an end-to-end validation of the DAWN project output. Thus, we plan
regular hackathons and contact points (via students and collaborators)
at multiple levels of the stack, from interfaces to pipelines to
substrates.

We believe the goals and research questions we have outlined here are
of broad technical merit to the software systems and computer
architecture communities and form a promising roadmap for future
data-intensive systems research at large. Our early results deploying
systems including Snorkel, MacroBase, and DeepDive in production have
confirmed our belief in the opportunity represented by the DAWN
project, and, looking forward, even incremental progress towards these
goals promises to radically improve upon the state of the art.\vspace{-.5em}

\begin{center}
\noindent\rule{8cm}{0.4pt}
\end{center}
\vspace{-1.5em}

\bibliography{dawn} \bibliographystyle{abbrv}

\end{document}